\documentclass[sigconf]{acmart}
\AtBeginDocument{%
  }

\copyrightyear{2025}
\acmYear{2025}
\setcopyright{cc}
\setcctype{by}
\acmConference[CIKM '25]{Proceedings of the 34th ACM International Conference on Information and Knowledge Management}{November 10--14, 2025}{Seoul, Republic of Korea}
\acmBooktitle{Proceedings of the 34th ACM International Conference on Information and Knowledge Management (CIKM '25), November 10--14, 2025, Seoul, Republic of Korea}
\acmDOI{10.1145/3746252.3761394}
\acmISBN{979-8-4007-2040-6/2025/11}

\usepackage{subfigure}
\usepackage{amsmath}
\usepackage{amsthm}
\usepackage{bm}
\usepackage{booktabs}
\usepackage{multirow}
\usepackage{xcolor}
\usepackage{float}
\newcommand{\m}{\textsc{SST}}
\newcommand{\mf}{Mambaformer}

\definecolor{deepred}{RGB}{152, 1, 0}

\begin{document}

\title{{{\m}: Multi-Scale Hybrid Mamba-Transformer Experts for Time Series Forecasting}}

\author{Xiongxiao Xu}
\affiliation{%
  \institution{Illinois Institute of Technology}
  \city{Chicago}
  \state{IL}
  \country{USA}
}
\email{xxu85@hawk.illinoistech.edu}

\author{Canyu Chen}
\affiliation{%
  \institution{Illinois Institute of Technology}
  \city{Chicago}
  \state{IL}
  \country{USA}
}
\email{cchen151@hawk.illinoistech.edu}

\author{Yueqing Liang}
\affiliation{%
  \institution{Illinois Institute of Technology}
  \city{Chicago}
  \state{IL}
  \country{USA}
}
\email{yliang40@hawk.illinoistech.edu}

\author{Baixiang Huang}
\affiliation{%
  \institution{Emory University}
  \city{Atlanta}
  \state{GA}
  \country{USA}
}
\email{baixiang.huang@emory.edu}

\author{Guangji Bai}
\affiliation{%
  \institution{Emory University}
  \city{Atlanta}
  \state{GA}
  \country{USA}
}
\email{gbai5@emory.edu}

\author{Liang Zhao}
\affiliation{%
  \institution{Emory University}
  \city{Atlanta}
  \state{GA}
  \country{USA}
}
\email{liang.zhao@emory.edu}

\author{Kai Shu}
\affiliation{%
  \institution{Emory University}
  \city{Atlanta}
  \state{GA}
  \country{USA}
}
\email{kai.shu@emory.edu}

\renewcommand{\shortauthors}{Xiongxiao Xu et al.}

\begin{abstract}
Time series forecasting has made significant advances, including with Transformer-based models. The attention mechanism in Transformer effectively captures temporal dependencies by attending to all past inputs simultaneously. However, its quadratic computational complexity with respect to sequence length limits the scalability for long-range modeling. Recent state space models (SSMs) such as Mamba offer a promising alternative by achieving linear complexity without attention. Yet, Mamba compresses historical information into a fixed-size latent state, potentially causing information loss and limiting representational effectiveness. This raises a key research question: \textit{Can we design a hybrid Mamba-Transformer architecture that is both effective and efficient for time series forecasting?} To address it, we adapt a hybrid Mamba-Transformer architecture \textit{{\mf}}, originally proposed for language modeling, to the time series domain. Preliminary experiments reveal that naively stacking Mamba and Transformer layers in {\mf} is suboptimal for time series forecasting, due to an \textit{information interference} problem. To mitigate this issue, we introduce a new time series decomposition strategy that separates time series into long-range patterns and short-range variations. Then we show that Mamba excels at capturing long-term structures, while Transformer is more effective at modeling short-term dynamics. Building on this insight, we propose \textit{State Space Transformer ({\m})}, a multi-scale hybrid model with expert modules: a Mamba expert for long-range patterns and a Transformer expert for short-term variations. To facilitate learning the patterns and variations, {\m} employs a multi-scale patching mechanism to adaptively adjust time series resolution: low resolution for long-term patterns and high resolution for short-term variations. Comprehensive experiments on real-world datasets demonstrate that {\m} achieves state-of-the-art performance while scaling linearly with sequence length ($O(L)$). The code is available here\footnote{https://github.com/XiongxiaoXu/SST}.
\end{abstract}



\maketitle

\begin{figure}[t]
\centering
\includegraphics[width=0.47\textwidth]{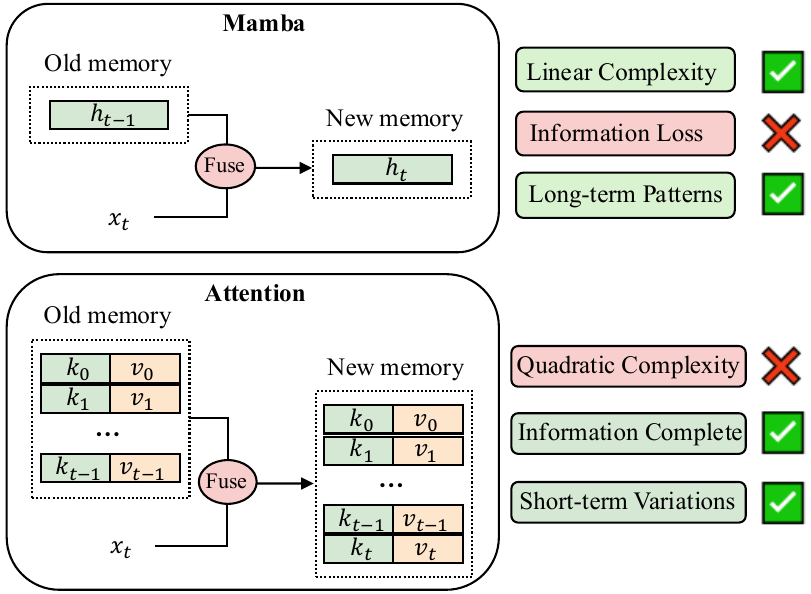}
\caption{Comparison of memory mechanism between Mamba and attention, where $x_i$ denotes the input token of $i-th$ step. Top: Mamba, as a RNN-like mechanism, compresses previous tokens into a fixed-size state $h_{t-1}$, which serves as the memory. When the current token $x_t$ occurs, $x_t$ is incorporated into $h_{t-1}$, leading to a new memory $h_t$ with the same size. This fixed size means that the memory is inherently lossy but linearly efficient. Bottom: Attention stores all previous tokens’ keys $k$ and values $v$ as memory. The memory is updated by continuously adding the current token’s key and value, so the memory is lossless. Therefore, attention can effectively manage short sequences but may encounter computational difficulties with longer ones.}
\vspace{-0.3cm}
\label{fig:mamba-att}
\end{figure}

\section{Introduction}
Time series forecasting has long served as a foundational task in a wide range of real-world applications~\cite{xu2025beyond}, including weather forecasting~\cite{zhu2023weather2k}, stock prediction~\cite{chen2023chatgpt}, and scientific computing~\cite{xu2023machine}. For example, in high-performance computing (HPC), ML-based surrogate models are trained to accelerate large-scale simulations of supercomputers by predicting their billion or even trillion behaviors over various timescales~\cite{cruz2023hybrid}, thereby avoiding a huge amount of energy consumption of supercomputers~\cite{xu2024online}.

Transformer-based and large language model (LLM)-based forecasters~\cite{nie2023a,jin2023time}, which are pre-trained on Transformer architectures, have demonstrated superiority in time series forecasting. The core strength of Transformer models lies in their attention mechanism, which identifies the most relevant information across the entire input sequence, and captures complex time series dependencies. However, the quadratic computation complexity of the attention limits its scalability for long-range sequence modeling. Recent advancements in state space models (SSMs), such as Mamba~\cite{gu2023mamba}, attempt to address this limitation by introducing linear-complexity through integrating gating, convolutions, and input-dependent token selection. While Mamba reduces the computational cost of the attention, it memorizes information by compressing the past information into a fixed-size latent state, which can result in information loss and constrained performance. Moreover, recent findings show that SSMs and Transformers are complementary for language modeling~\cite{lieber2024jamba,park2024can,fathi2023block}. Motivated by it, we pose a research question: \textit{Can we design a hybrid Mamba-Transformer architecture that is both effective and efficient for time series forecasting?}

However, it presents several significant challenges for hybrid Mamba-Transformer model in time series forecasting. First, while Mamba and Transformer each have distinct strengths, i.e., Mamba offers efficiency, and Transformer excels at expressivity, their limitations are also complementary. It remains unclear how to design a hybrid model that leverages their respective advantages without inheriting their drawbacks. Second, both models originate from language modeling, and how to embed time series–specific inductive biases into a hybrid architecture remains an open question. A straightforward approach is to adapt \textit{{\mf}}~\cite{park2024can}, which is a hybrid Mamba-Transformer framework initially developed in language modeling. {\mf} achieves strong performance in text modeling by stacking Mamba layers with attention blocks. Given the sequential nature shared by language and time series data, one might expect {\mf} to generalize reasonably well to time series forecasting. However, our preliminary results suggest that the naively interleaving of Mamba and Transformer layers in {\mf} even fails to outperform a simple linear baseline \textit{DLinear}\cite{zeng2023transformers}. We attribute this degradation to an \textit{information interference} issue where Mamba and attention mechanisms conflict in how they encode and retrieve temporal information.

To address this issue, we introduce a new time series decomposition method that breaks a sequence into long-range patterns and short-range variations, which can be modeled independently by Mamba and Transformer, respectively. This strategy decouples the Mamba-Transformer stack, thus avoiding the \textit{information interference} issue. Mamba, which maintains knowledge by a fixed-size latent state inherited from recurrent neural networks (RNNs)~\cite{medsker2001recurrent} but may lose information, is well-suited for capturing long-term patterns with filtering out variations in the long run. In contrast, Transformer, which does not compress context into a smaller state but is prohibitively computing expensive, is more effective in capturing short-term variations. To harness their complementary strengths, we propose a multi-scale hybrid Mamba-Transformer experts model \textit{State Space Transformer ({\m})}. 

\begin{figure}[t]
\centering
\includegraphics[width=0.47\textwidth]{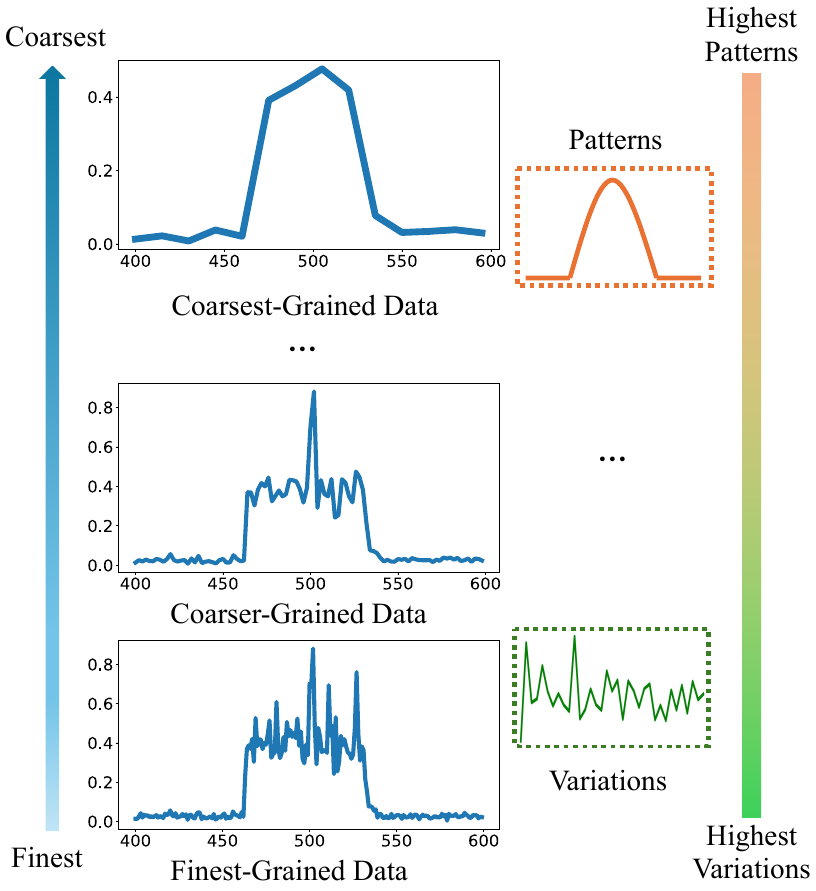}
\caption{From bottom to top, as time series resolution shifts from fine-grained to coarse-grained, patterns become increasingly pronounced while variations diminish.}
\vspace{-0.6cm}
\label{fig:resolution}
\end{figure}

The core idea of {\m} is to separately model patterns in long range by Mamba and variations in short range by Transformer. Specifically, {\m} includes three components: (a) multi-scale patcher, (b) hybrid Mamba-Transformer experts, and (c) long-short router. (a) The multi-scale patcher modifies resolutions of patched time series (PTS)~\cite{nie2023a}: low-resolution for long range, and high-resolution for short range. Figure~\ref{fig:resolution} illustrates that patterns are more discernible at a coarser granularity, while variations emerge at a finer granularity. Therefore, we employ larger patches and longer stride for long range to obtain a low-resolution PTS; smaller patches and shorter stride on short term leads to a high-resolution PTS. Moreover, since existing literature lacks a quantitative metric to assess the resolution of PTS, we propose a metric \textit{$R_{PTS}$} to precisely quantify the granularity of PTS. (b) Inspired by the idea of Mixture-of-Experts (MoEs)~\cite{fedus2022switch}, {\m} adopts Mamba as an expert to model patterns in long range and Transformer as the other expert to capture variations in short range. Unlike traditional MoEs' approaches where roles of experts are ambiguous, we clearly delineates the responsibilities for two experts. (c) To adaptively integrate the two specialized experts, the long-short router is proposed to learn their contributions, thus leading to enhanced forecasting performance. Remarkably, {\m} achieves SOTA performance and also ensures a linear complexity $O(L)$ on input length $L$. To the best of our knowledge, {\m} is the first hybrid Mamba-Transformer architecture in time series forecasting. We summarize our contributions as follows:
\begin{itemize}
    \item We investigate the potential of hybrid Mamba-Transformer architectures for time series forecasting by adapting {\mf} to this domain. Our analysis reveals that naively stacking Mamba and Transformer layers, as done in {\mf}, leads to an information interference issue, resulting in suboptimal forecasting performance.
    \item We introduce a new time series decomposition method that separates input sequences into long-range patterns and short-range variations, enabling Mamba and Transformer modules to function independently, thus eliminating information interference issue inherent in naive hybrid architectures.
    \item We propose a novel hybrid Mamba-Transformer experts architecture {\m} that leverages Mamba as a long-term patterns expert and Transformer as a short-term variations expert. {\m} also incorporates a multi-scale patching mechanism to adjust time series resolution and a long-short routing module to adaptively fuse the expert outputs. Extensive experiments show that {\m} achieves state-of-the-art forecasting performance with linear complexity $O(L)$ in sequence length $L$.
\end{itemize}

\section{Preliminaries}\label{sec:preli}
\subsection{Problem Statement}
In time series forecasting, historical time series with a look-back window $\mathcal{L}=(\mathbf{x}_1,\mathbf{x}_2,..,\mathbf{x}_{L})\in\mathbb{R}^{L\times M}$ of length $L$ is given, where each $\mathbf{x}_t\in\mathbb{R}^M$ at time step $t$ is with $M$ variates, we aim to forecast $F$ future values $\mathcal{F}=(\mathbf{x}_{L+1},\mathbf{x}_{L+2},..,\mathbf{x}_{L+F})\in\mathbb{R}^{F\times M}$ with forecasting length $F$. Long-range time series $\mathcal{L}$ denotes the full range of look-back window $\mathcal{L[:]}$, and short-range time series $\mathcal{S}\in\mathbb{R}^{S\times M}$ denotes the partial latest range $\mathcal{L}[-S:]$, $S<L$. Input $\mathcal{L}$ and output $\mathcal{F}$ can also be denoted as $(\mathbf{x}^{(1)},\mathbf{x}^{(2)},..,\mathbf{x}^{(M)})$ indicating $M$ dimensions.

\subsection{State Space Models}
The State Space Model (SSM) is a class of sequence modeling frameworks that are broadly related to RNNs, and CNNs, and classical state space models~\cite{gu2021combining}.
They are inspired by a continuous system that maps an input function $x(t)\in\mathbb{R}$ to an output function $y(t)\in\mathbb{R}$ through an implicit latent state $h(t)\in\mathbb{R}^{N}$ as follows:
\begin{align}
    h'(t)=\mathbf{A}h(t)+\mathbf{B}x(t),  y(t)=\mathbf{C}h(t)
\end{align}
where $\mathbf{A}\in\mathbb{R}^{N\times N}$, $\mathbf{B}\in\mathbb{R}^{N\times 1}$, and $\mathbf{C}\in\mathbb{R}^{1\times N}$ are learnable matrices. SSM can be discretized from continuous signals into discrete sequences by a step size $\Delta$ as follows:
\begin{align}\label{eq2}
    h_t=\overline{\mathbf{A}}h_{t-1}+\overline{\mathbf{B}}x_t, y=\mathbf{C}h_t
\end{align}
The discrete parameters ($\overline{\mathbf{A}}$, $\overline{\mathbf{B}}$) can be obtained from continuous parameters ($\Delta$, $\mathbf{A}$, $\mathbf{B}$) through a discretization rule, such as zero-order hold (ZOH) rule $\overline{\mathbf{A}}=\exp(\Delta\mathbf{A})$, $\overline{\mathbf{B}}=\exp(\Delta\mathbf{A})^{-1}(\exp(\Delta\mathbf{A})-\mathbf{I})\cdot\Delta\mathbf{B}$. After discretization, the model can be computed in two ways, either as a linear recurrence for inference as shown in Equation~\ref{eq2}, or as a global convolution for training as the following Equation~\ref{eq3} shows, where $\overline{\textbf{K}}$ is a convolution kernel:
\begin{align}\label{eq3}
    \overline{\textbf{K}} = (\textbf{C}\overline{\textbf{B}}, \textbf{C}\overline{\textbf{A}\textbf{B}},..., \textbf{C}\overline{\textbf{A}}^k\overline{\textbf{B}},...), y= x\ast\overline{\textbf{K}}
\end{align}
S4 is a structured SSM where the specialized Hippo~\cite{gu2020hippo} structure is imposed on the matrix $\mathbf{A}$ to capture long-range dependency. Building upon S4, Mamba~\cite{gu2023mamba} incorporates a selective SSM to propagate or forget information along the sequence, and a hardware-aware algorithm for efficient implementation. 

\subsection{Transformer} Transformer~\cite{vaswani2017attention} is becoming a basis architecture for deep learning. The core module of Transformer is the multi-head attention. Each head in the layer simultaneously transforms input into query matrices $Q$, key matrices $K$, and value matrices $V$. The output is based on the scaled dot-product:
\begin{equation}
    Attention(\mathbf{Q}, \mathbf{K}, \mathbf{V})=softmax(\frac{QK^T}{\sqrt{d_k}})V
\end{equation}
where $\sqrt{d_k}$ is a scaling factor to avoid small gradients. The attention mechanism has demonstrated effectiveness in a wide range of applications; however, the quadratic complexity poses a significant challenge for scaling to long input.

\subsection{Time Series Decomposition}
Time series decomposition is a classical technique that separates a time series into multiple interpretable components. One widely used method is STL (Seasonal-Trend decomposition using LOESS)~\cite{cleveland1990stl}, which decomposes a time series into three additive components: trend, seasonal, and residual (noise). This separation enables models to focus on distinct temporal dynamics of different scales. Formally, a univariate time series $\mathbf{x}$ (ignore superscript) can be expressed as:
\begin{equation}
    \mathbf{x} = \mathbf{T} + \mathbf{S} + \mathbf{R}
\end{equation}
where $\mathbf{T}$ captures the long-term trend, $\mathbf{S}$ represents seasonal or periodic patterns, and $\mathbf{R}$ denotes the residual fluctuations. STL decomposition is commonly used in time series literature~\cite{zhou2022fedformer,theodosiou2011forecasting}, where understanding and isolating periodicity or trends is critical for accurate forecasting and decision-making.

\begin{figure*}[t]
    \centering
    \includegraphics[width=\textwidth]{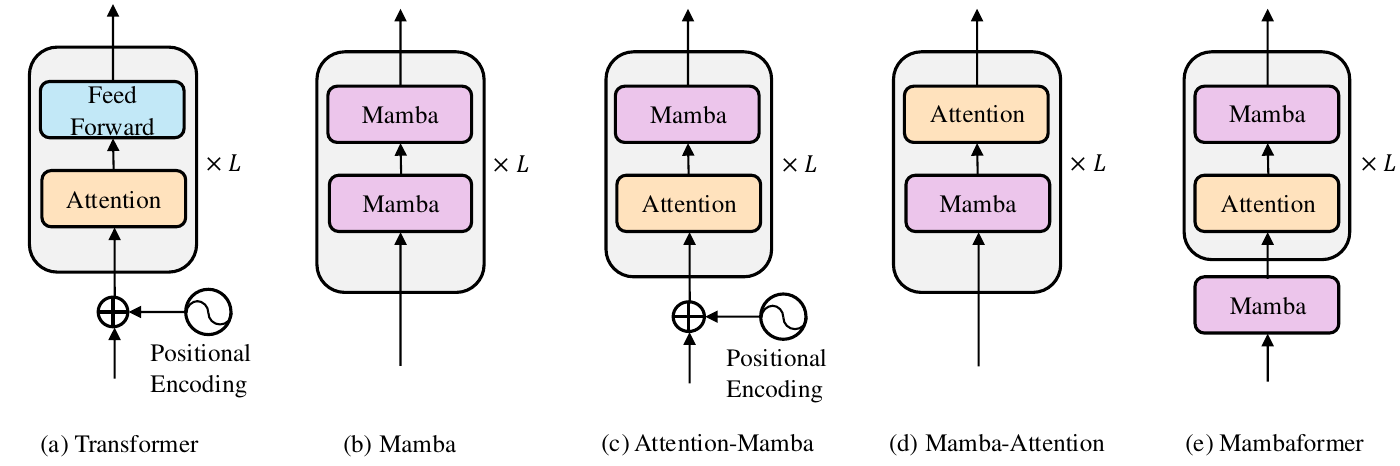}
     \caption{The architectures of {\mf} family. Positional encoding is optional across these variants. Mamba layers inherently encode positional information by state dynamics while Transformer layers require explicit positional encoding. When a Mamba layer is before an attention layer (Mamba, Mamba-Attention, and Mambaformer), positional encoding can be omitted. However, if the attention layer comes first (Transformer and Attention-Mamba), positional encoding is necessary.
     }
    \label{fig:mffamily}
\end{figure*}

\section{Is Vanilla Stacking of Mamba-Transformer Effective for Time Series Forecasting?}\label{sec:vanilla_stack}
In this section, we evaluate the effectiveness of the hybrid \textit{{\mf}} architecture in time series, which is originally proposed for language modeling. Our experiments reveal that a straightforward stacking of Mamba and Transformer layers struggles to model temporal dependencies in time series data, potentially due to an \textit{information interference} issue. To address it, we propose to decompose time series into long-range patterns and short-range variations, which can be addressed separately by Mamba and Transformer.

\subsection{Adapting {\mf} to Time Series}
We adapt the {\mf} architecture~\cite{park2024can} to time series forecasting. It attempts to combine strengths of Mamba and attention by interleaving their layers. To make {\mf} compatible with time series data, we introduce two types of embedding layers:
\begin{itemize}
    \item \textbf{Conv}: A one-dimensional convolutional encoder that projects raw time series into a high-dimensional space while preserving local semantics~\cite{chang2023llm4ts}.
    \item \textbf{PI (Patch + Instance Normalization)}: We employ a time series encoding methods in PatchTST~\cite{nie2023a}. It utilizes a patching strategy to reduce sequence length and preserve local structure, coupled with instance normalization~\cite{ulyanov2016instance,kim2021reversible} to reduce distribution shift between training and testing.
\end{itemize}
The embedding layer is followed by the core {\mf} architecture (Figure~\ref{fig:mffamily}) to learn complex temporal dependencies. Finally, a linear layer maps the high-dimensional embeddings back to the original time series dimension to obtain forecasting results. 
\begin{table}[t]
    \centering
    \scalebox{0.82}{
    \begin{tabular}{lcccccc}
        \toprule
        Methods & \multicolumn{2}{c}{\textbf{ETTh1}} & \multicolumn{2}{c}{\textbf{Weather}} & \multicolumn{2}{c}{\textbf{Traffic}} \\
        \cmidrule(r){2-3} \cmidrule(r){4-5} \cmidrule(r){6-7}
        Metrics & MSE & MAE & MSE & MAE & MSE & MAE \\
        \midrule
        DLinear & \textbf{0.455} & \textbf{0.451} & \textbf{0.265} & \textbf{0.316} & \textbf{0.624} & 0.383 \\ \hline
        Transformer (Conv) & 0.991 & 0.790 & 0.491 & 0.485 & 0.672 & 0.397 \\
        Transformer (PI) & 0.652 & 0.731 & 0.314 & 0.467 & 0.648 & 0.385 \\
        Mamba (Conv) & 1.417 & 0.914 & 0.890 & 0.677 & 1.097 & 0.558 \\
        Mamba (PI) & 1.172 & 0.662 & 0.561 & 0.419 & 0.965 & 0.415 \\ 
        Attention-Mamba (Conv) & 0.995 & 0.792 & 0.656 & 0.591 & 0.661 & 0.394 \\
        Attention-Mamba (PI) & 0.735 & 0.659 & 0.482 & 0.463 & 0.629 & \textbf{0.358} \\ 
        Mamba-Attention (Conv) & 0.973 & 0.727 & 0.798 & 0.671 & 0.935 & 0.488 \\
        Mamba-Attention (PI) & 0.741 & 0.663 & 0.569 & 0.454 & 0.843 & 0.441 \\ 
        Mambaformer (Conv) & 0.962 & 0.721 & 0.842 & 0.679 & 0.733 & 0.401 \\
        Mambaformer (PI) & 0.693 & 0.645 & 0.527 & 0.432 & 0.681 & 0.374 \\
        \bottomrule
    \end{tabular}
    }
    \caption{Performance comparison between DLinear and {\mf} family. The best performance is highlighted in bold. It shows that {\mf} family cannot outperform DLinear, suggesting that the vanilla stacking of Mamba and Transformer is not effective for time series forecasting.}
    \label{tab:mffamily}
    \vspace{-0.9cm}
\end{table}
Figure~\ref{fig:mffamily} illustrates five architectural variants of {\mf}, including the original Transformer and Mamba, as well as hybrid combinations (Attention-Mamba, Mamba-Attention, and {\mf}). We call them Mambaformer family in this work.

\subsection{{\mf} is Ineffective for Time Series}
We compare {\mf} family against an embarrassingly simple linear model \textit{DLinear}~\cite{zeng2023transformers}. Table~\ref{tab:mffamily} presents the experimental results, where the look-back window is fixed at $L=196$, and performance is averaged across multiple forecasting horizons $F \in \{96, 192, 336, 720\}$. The results indicate that all {\mf} family variants perform notably worse than the simple baseline DLinear. For example, on the ETTh1 dataset, DLinear achieves an MSE of 0.455, whereas the best-performing {\mf} variant Mambaformer (PI) reaches 0.693. It suggests that naively stacking Mamba and Transformer layers is ineffective in time series forecasting, without accounting for the uniqueness of time series data.

We attribute the poor performance of {\mf} family to \textit{information interference} issue in time series, where Mamba and attention mechanisms conflict in how they encode and retrieve temporal information. Specifically, Mamba compresses historical inputs into a fixed-size hidden state, inevitably discarding some information. When Transformer’s attention mechanism is applied to this lossy representation, its effectiveness is diminished, since it can no longer attend to fine-grained details. Such degradation may be tolerable in text, where semantically meaningful tokens provide redundancy. However, time series data lacks such inherent structure, making it more sensitive to information loss. Therefore, a more principled and task-specific integration of sequential state space modeling and attention mechanisms is needed to fully exploit their potential in time series forecasting.

\subsection{Mamba for Long-Range Patterns and Transformer for Short-Range Variations}
The issue of \textit{information interference} arises from the naive stacking of Transformer and Mamba architectures. To address this, we propose a new time series strategy to separately model long-term patterns and short-term variations, which can be addressed by Mamba and Transformer independently, thus avoiding their conflicts.

\begin{figure}[t]
    \centering
    \includegraphics[width=0.48\textwidth]{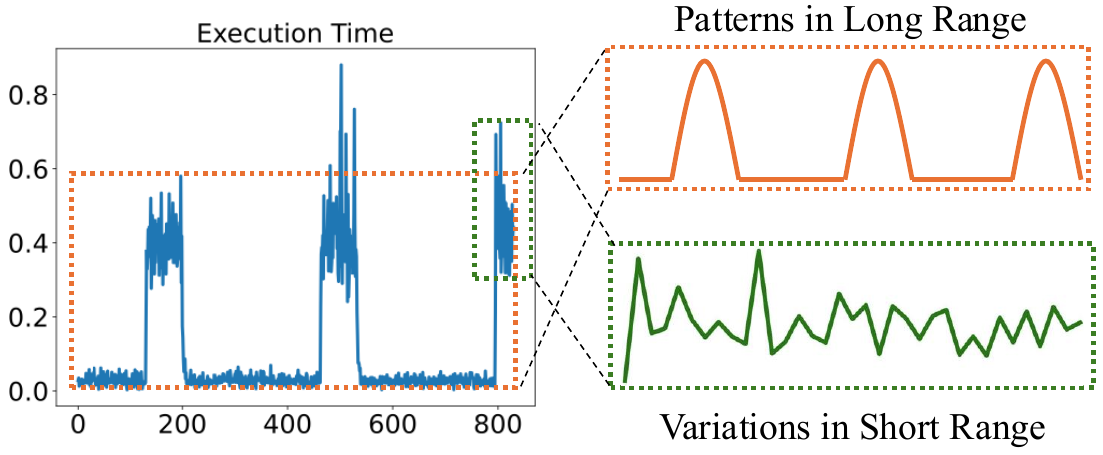}
     \vspace{-0.5cm}
     \caption{A real-world HPC dataset reflecting supercomputers' behaviors can be decomposed into patterns and variations by ranges. Long-term patterns (orange line) indicate repeated up-and-down trends because supercomputers intermittently executes applications, and short-term variations (green line) are extreme execution times caused by sudden network congestion.}
     \vspace{-0.5cm}
    \label{fig:decomposition}
\end{figure}

We first describe the decomposition technique with a real-world HPC dataset~\cite{xu2023machine} which describes execution times of supercomputers across different simulation iterations in the scientific computing domain. Figure~\ref{fig:decomposition} shows that time series can naturally be decomposed into long-range patterns and short-range variations. In this scenario, long-term patterns represent repeated up-and-down trends as supercomputers intermittently executes applications, and short-term variations are extreme execution times caused by sudden network congestion. For long-term forecasting, the model should focus on overall patterns while ignoring local deviations such as outliers. In contrast, short-term forecasting benefits from emphasizing local fluctuations, where patterns are less salient. For example, occasional spikes due to network congestion represent anomalies within the broader trend but are critical for predicting the immediate next-term system behavior~\cite{xu2024surrogate}.

Note that our decomposition is actually a variant of the classical STL decomposition. Unlike STL, which treats trend, seasonality, and residuals as symmetric components, our approach explicitly reflects the asymmetry between long- and short-range dependencies. Formally, we reinterpret STL as $\mathbf{x} = \mathbf{T} + \mathbf{S} + \mathbf{R} = (\mathbf{T}_{Long} + \mathbf{S}_{Long}) + \mathbf{R}_{Long} \approx \mathbf{Patterns}_{Long} + \mathbf{Variations}_{Short}$, where long-range patterns encompass both trend and seasonality, and short-range variations represent noise or high-frequency signals that are contextually meaningful only in the short term.

As discussed in Figure~\ref{fig:mamba-att}, Mamba and Transformer differ fundamentally in memory mechanisms and temporal modeling. Mamba scales linearly with input length but inherently compresses information, making it a desirable property for capturing salient long-range patterns while ignoring noise. On the other hand, Transformer retains all historical inputs through key-value attention, enabling fine-grained modeling of short-term variations. Built on these complementary strengths, we propose a design principle for hybrid architectures: Mamba for long-range pattern extraction and Transformer for short-range variation modeling. 

\section{Methodology}\label{sec:method}
\begin{figure*}[!t]
    \centering
    \includegraphics[width=0.99\textwidth]{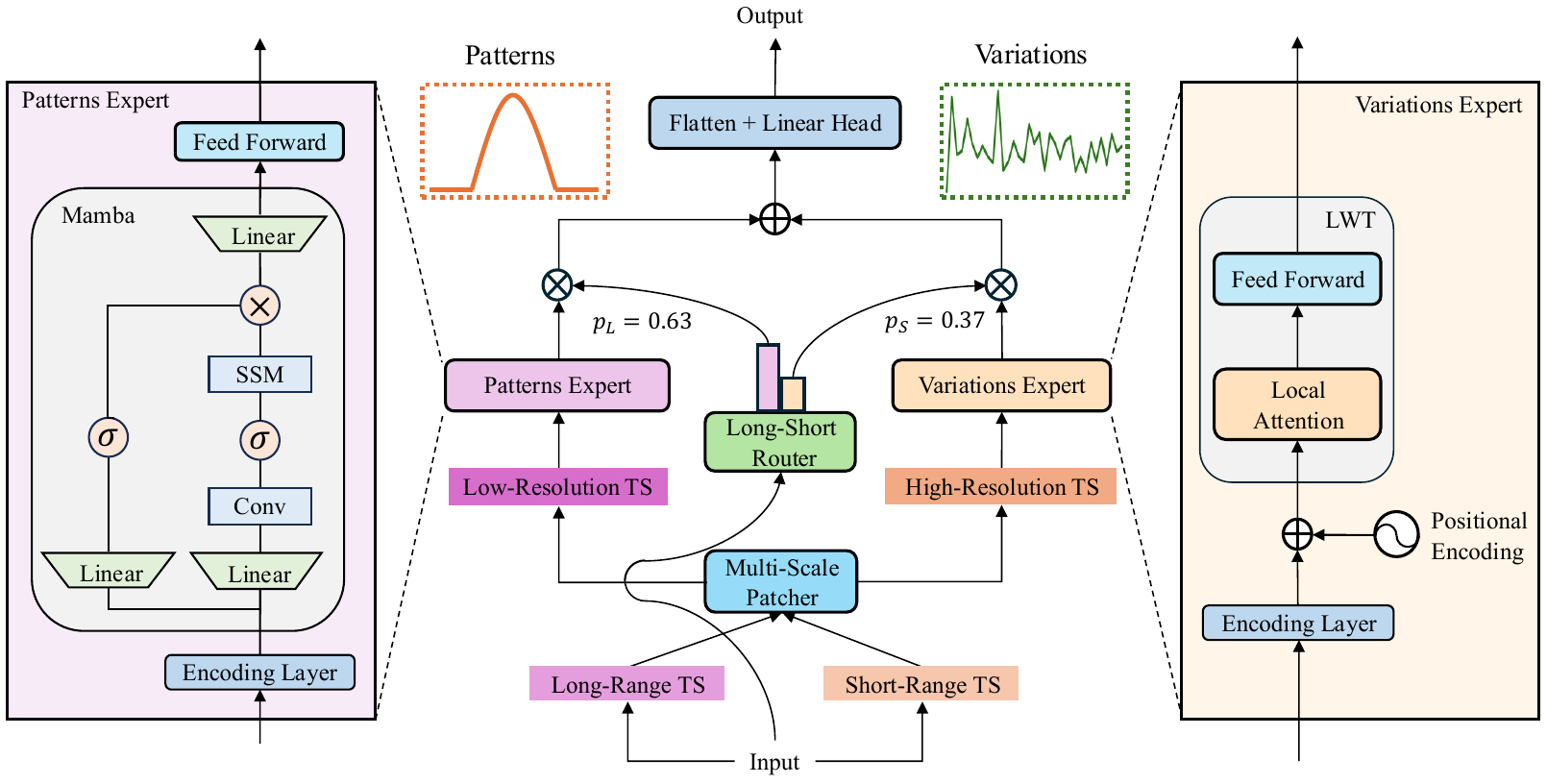}
     \caption{The overview of the {\m}. The multi-scale patcher transforms input time series in different resolutions according to ranges. The Mamba is dedicated for long-term patterns and the LWT is responsible for short-term variations. The long-short router adaptively learns the contributions of the two experts.}
    \label{fig:framework}
\end{figure*}

In this section, we introduce the proposed multi-scale hybrid architecture \textit{(State Space Transformer) {\m}}. As shown in Figure~\ref{fig:framework}, {\m} primarily includes three modules: multi-scale patcher, hybrid Mamba-Transformer experts, and long-short router. Multi-scale patcher transforms input time series into distinct granularities according to ranges. Mamba extracts patterns from coarse-grained long-range series, and Transformer captures variations from fine-grained short-range series. The long-short router dynamically learns contributions of the two experts, finally leading to enhanced performance.

\subsection{Multi-Scale Resolutions}
\textbf{Patching.}
Figure~\ref{fig:resolution} suggests that time series patterns emerge when viewed at a broader-scale granularity, while variations become clearer when examined at a finer-scale granularity. Motivated by this, we propose a multi-scale patcher to modify resolutions of long- and short-range time series, i.e., low resolution for long range and high resolution for short range. The patching technique~\cite{nie2023a} is increasingly popular in time series analysis as it can retain local semantic information and reduce computation overhead by aggregating time steps into subseries-level patches. We modify time series resolution by adjusting the patch and stride length. Formally, input time series $\mathcal{L}$ is divided into $M$ univariate time series $\mathcal{L}=(\mathbf{x}^{(1)},\mathbf{x}^{(2)},..,\mathbf{x}^{(M)})\in\mathbb{R}^{L\times M}, \mathbf{x}^{(i)}\in\mathbb{R}^{L\times 1}$. The patching process involves two factors: the patch length $P$ and the stride length $Str$ (the interval between the end point of two consecutive patches). Accordingly, the number of patches is $N=\lfloor\frac{L-P}{Str}\rfloor+1$, and the patcher for each variate $\mathbf{x}^{(i)}$ generates a sequence of patches $\mathbf{x}_p^{(i)}\in\mathbb{R}^{N\times P}$, named patches time series (PTS). Note that for unpatched time series $\mathbf{x}^{(i)}\in\mathbb{R}^{L\times 1}$, $L$ is the sequence dimension and $1$ denotes the variate dimension. Correspondingly, for PTS $\mathbf{x}_p^{(i)}\in\mathbb{R}^{N\times P}$, $N$ is the new sequence dimension and $P$ is the new variate dimension.
\\\textbf{Resolution.}
Intuitively, larger patches $P$, longer stride $Str$, and accordingly less number of patches $N$ indicates a low resolution. We adopt such setting to generate a low-resolution time series for long range. It allows {\m} focus on modeling long-term patterns and ignoring small fluctuation. By contrast, smaller patches and shorter stride imply a high resolution. It enables {\m} to depict short-term nuances. Although the patching is becoming popular in time series community, no existing work tries to quantify resolutions of PTS. To mitigate the gap, we define a new metric as follows:
\begin{definition}
   PTS Resolution. Let $P$, $Str$, and $N$ denote patch length, stride length, and number of patches for a patched time series. The resolution of a PTS is defined as $R_{PTS}=\frac{\sqrt{P}}{Str}$.
\end{definition}
The definition of PTS resolution takes two factors into account. First, PTS resolution aims to quantify the relative granularity of a PTS compared to its unpatched one. Second, PTS resolution describes the amount of temporal information of a PTS, in both the sequence dimension and variate dimension. Similar to the definition of the image resolution~\cite{boellaard2004effects} equaling to $|width|\times |height|$, we multiply the two dimensions of PTS, but with a penalized root at variate dimension, i.e., $|sequence| \times \sqrt{|variate|}$ as we focus more on the sequence (temporal) dimension of PTS. For example, after an univariate time series $\mathbf{x}^{(i)}\in\mathbb{R}^{L\times 1}$ is patched into a PTS $\mathbf{x}_p^{(i)}\in\mathbb{R}^{N\times P}$, 
the resolution $R_{PTS}$ is defined as:
\begin{equation}
R_{PTS}=N\sqrt{P}=(\lfloor\frac{L-P}{Str}\rfloor+1)\sqrt{P}\approx\frac{\sqrt{P}}{Str}
\end{equation}
In the last approximation, $L$ is ignored as $R_{PTS}$ aims to quantify the relative resolution.

\subsection{Hybrid Mamba-Transformer Experts}
\begin{figure}[t]
\centering
\includegraphics[width=0.45\textwidth]{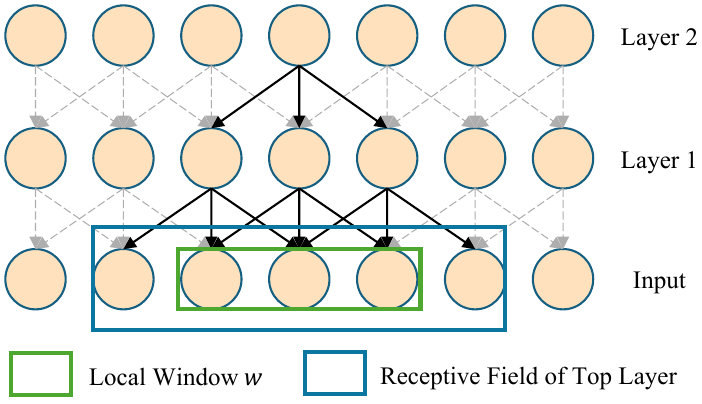}
\caption{LWT employs a fixed-size window $w$ to force each token only attend to local tokens within the window. By stacking multiple layers, the upper attention layer $l$ can aggregate information from the lower layer and obtain a large receptive field $l\times w$. The complexity of Transformer can be reduced from quadratic to linear with the local window.}
\label{fig:lwt}
\end{figure}

Inspired by Mixture-of-Experts (MoEs)~\cite{jacobs1991adaptive,Shazeer2017Outrageously,fedus2022switch}, we introduce a hybrid Mamba-Transformer experts architecture. Unlike ambiguous roles in traditional MoEs models, we assign clear patterns and variations roles to the two experts.
\\\textbf{Patterns Expert.} Mamba achieves impressive performance on tasks requiring scaling to long-range sequences as it introduces a selective mechanism to remember relevant patterns and filter out irrelevant noises indefinitely. 
Consequently, we incorporate the Mamba block as a expert to extract long-term patterns and filter out small variations in long-range time series. As shown in Figure~\ref{fig:framework}, the patterns expert encodes long-range PTS $\mathbf{x}_{pL}^{(i)}\in\mathbb{R}^{N_L\times P_L}$ into high-dimension space $\mathbf{x}_L^{(i)}\in\mathbb{R}^{N_L\times D}$ in the encoding layer, and a Mamba block is followed.
Mamba takes an input $\mathbf{x}_L^{(i)}$ and expands the dimension by two input linear projections. For one projection, Mamba processes the expanded embedding through a convolution and a SiLU activation before feeding into the SSM. The core discretized SSM module is able to select input-dependent patterns and filter out irrelevant variations. The other projection followed by SiLU activation, as a residual connection, is combined with output of the SSM module through a multiplicative gate. Finally, Mamba delivers output $\mathbf{z}_L^{(i)}\in\mathbb{R}^{N_L\times D}$ through an output linear projection. Note that we do not need a positional embedding typically existing in Transformer as Mamba's recurrence mechanism~\cite{gu2021combining} inherently encodes positions.
\\\textbf{Variations Expert.} Transformer is potential to capture variations as discussed in the Section~\ref{sec:vanilla_stack}. However, it lacks locality inductive bias and quadratic complexity further impacts its adoption. Therefore, we propose a local window Transformer (LWT) with enhanced locality-awareness capabilities to capture short-term variations in short range. Figure~\ref{fig:framework} shows that the variations expert first projects short-range PTS $\mathbf{x}_{pS}^{(i)}$ and positional information into a embedding $\mathbf{x}_S^{(i)}\in\mathbb{R}^{N_S\times D}$. The embedding is fed into LWT where the local window attention forces each token only care surrounding tokens within the window. We denote the output of LWT as $\mathbf{z}_S^{(i)}\in\mathbb{R}^{N_S\times D}$. Figure~\ref{fig:lwt} illustrates the mechanism of local window attention. Despite its strong local inductive bias, LWT maintains an extensive receptive field. By stacking multiple layers, the upper layers gain access to all input locations, enabling the construction of representations that integrate information across the entire input, similar to the capabilities of CNNs~\cite{wu2019pay}. The LWT is able to increase the receptive field by stacking multiple layers, and the respective field of layer $l$ is $l\times w$. Moreover, the complexity reduces from $O(S^2)$ to $O(w*S)$ on the length of short-range time series (not considering the patching here).

\subsection{Mixture of Experts}
Inspired by the idea of MoE, we propose a long-short router to adaptively learn the contributions of the patterns expert and the variation expert.
\\\textbf{Long-Short Router.} In the MoE community, router~\cite{fedus2022switch} is a key module which primarily directs tokens to only a subset of experts for saving computational cost. Different from the path-assigning role of traditional MoE routers, the proposed long-short router is capable of learning the relative contributions of the two specialized experts, adaptively integrating long- and short-range time series representations.
Formally, the router projects input time series $\mathcal{L}\in \mathbb{R}^{L\times M}$ into the $D$-dimension space, subsequently transformed by a flatten layer into a vector $\mathbf{z}_R$.
A linear layer with a softmax function is followed to output two values $p_L, p_S \in (0,1)$, indicating two weights of patterns and variations.
\\\textbf{Forecasting Module.} The forecasting module flattens long-range embedding $\mathbf{z}_L^{(i)}$ and short-range embedding $\mathbf{z}_S^{(i)}$ into a single-row vector and concatenates them with respective weights $p_L$ and $p_S$. The resulting representation $\mathbf{z}_{LS}^{i}\in\mathbb{R}^{(N_S+N_L)LD}$ integrates long-term patterns and short-term variations. Finally, a linear head is employed to forecast $\mathbf{\hat{x}}^{(i)}=\{\hat{x}_{L+1},\hat{x}_{L+2},...,\hat{x}_{L+F}\}\in\mathbb{R}^{F\times1}$ for individual variate $i$.

\subsection{Linear Complexity Analysis.}  
The complexity of {\m} comes form two parts: Mamba and LWT. Mamba which adopts RNN-like mechanism achieves the linear complexity $O(L)$ on long-range time series $\mathcal{L}$. Given the attention operation occupies $O(w^2)$ complexity within a local window $w$ each time step, LWT consumes $O(w^2*\frac{S}{w})$ complexity on short-range time series $\mathcal{S}$. Considering the use of multi-scale patcher, the total complexity of {\m} is $O(\frac{L}{N_L}+\frac{wS}{N_S})$. Note that $w$, $N_S$, and $N_L$ are constant, and $S$ is linear related to and smaller than $L$.

\begin{table*}[t]
    \centering
    \caption{The statistics of seven real-world time series datasets.}
    \begin{tabular}{c|cccccccc}
    \toprule
        \textbf{Datasets} & ETTh1 & ETTh2 & ETTm1 & ETTm2 & Weather & ECL & Traffic  \\ \hline
        \textbf{Variates} & 7 & 7 & 7 & 7 & 21 & 321 & 862 \\
        \textbf{Timestamps} & 17,420 & 17,420 & 69,680 & 69,680 & 52,696 & 26,304 & 17,544 \\
    \bottomrule
    \end{tabular}\label{tab:dataset}
\end{table*}

\begin{table*}[t]
    \centering
    \caption{Multivariate time series forecasting results across seven datasets. The best and second-best performing models are highlighted in bold and underline.}
    \scalebox{0.78}{
    \begin{tabular}{c|ccccccccccccccccccccc}
    \toprule
        \multicolumn{2}{c}{} & \multicolumn{2}{c}{} & \multicolumn{6}{c}{Mamba-based} & \multicolumn{6}{c}{Transformer-based} & \multicolumn{6}{c}{MLP-based} \\
        \multicolumn{2}{c}{Models} & \multicolumn{2}{c}{SST} & \multicolumn{2}{c}{TimeMachine} & \multicolumn{2}{c}{S-Mamba} & \multicolumn{2}{c}{SiMBA} & \multicolumn{2}{c}{MTST} & \multicolumn{2}{c}{iTransformer} & \multicolumn{2}{c}{PatchTST} & \multicolumn{2}{c}{TimeMixer} & \multicolumn{2}{c}{RLinear} & \multicolumn{2}{c}{DLinear} \\ 
        \multicolumn{2}{c}{Metrics} & MSE & MAE & MSE & MAE & MSE & MAE & MSE & MAE & MSE & MAE & MSE & MAE & MSE & MAE & MSE & MAE & MSE & MAE & MSE & MAE \\ \hline 

        \parbox[t]{2mm}{\multirow{4}{*}{\rotatebox[origin=c]{90}{ETTh1}}} & 96 & \textbf{0.381} & 0.405 & 0.389 & 0.402 & 0.392 & \textbf{0.390} & 0.432 & 0.416 & \textbf{0.381} & 0.392 & \underline{0.386} & 0.405 & 0.414 & 0.419 & 0.403 & 0.421 & \underline{0.386} & \underline{0.395} & \underline{0.386} & 0.400 \\
        & 192 & \textbf{0.430} & 0.434 & \underline{0.435} & 0.440 & 0.449 & 0.439 & 0.451 & \textbf{0.422} & 0.437 & 0.438  & 0.441 & 0.436 & 0.460 & 0.445 & 0.451 & 0.447  & 0.437 & \underline{0.424} & 0.437 & 0.432 \\
        & 336 & \underline{0.443} & \textbf{0.446} & \underline{0.450} & \underline{0.448} & 0.467 & 0.481 & 0.477 & 0.465 & \textbf{0.425} & 0.466  & 0.487 & \underline{0.458} & 0.501 & 0.466 & 0.517 & 0.451  & 0.479 & \textbf{0.446} & 0.481 & 0.459 \\
        & 720 & 0.502 & 0.501 & \underline{0.480} & \textbf{0.465} & \textbf{0.475} & \underline{0.468} & 0.531 & 0.515 & 0.483 & 0.472  & 0.503 & 0.491 & 0.500 & 0.518 & 0.529 & 0.502 & 0.481 & 0.470 & 0.519 & 0.516 \\ \hline
        
        \parbox[t]{2mm}{\multirow{4}{*}{\rotatebox[origin=c]{90}{ETTh2}}} & 96 & 0.291 & \underline{0.346} & \textbf{0.230} & 0.349 & 0.292 & 0.357 & 0.338 & 0.370 & 0.304 & 0.359  & 0.297 & 0.349 & 0.302 & 0.348 & 0.316 & 0.360 & \underline{0.288} & \textbf{0.338} & 0.333 & 0.387 \\
        & 192 & \textbf{0.369} & 0.397 & \underline{0.371} & 0.400 & 0.380 & 0.402 & 0.410 & 0.465 & 0.381 & \underline{0.395}  & 0.380 & 0.400 & 0.388 & 0.400 & 0.402 & 0.415  & 0.374 & \textbf{0.390} & 0.477 & 0.476 \\
        & 336 & \textbf{0.374} & \textbf{0.414} & 0.402 & 0.449 & \underline{0.391} & \underline{0.420} & 0.481 & 0.441 & 0.399 & 0.417  & 0.428 & 0.432 & 0.426 & 0.433 & 0.417 & 0.433  & 0.415 & 0.426 & 0.594 & 0.541 \\
        & 720 & \textbf{0.419} & 0.447 & 0.425 & \underline{0.438} & 0.437 & 0.455 & 0.486 & 0.501 & 0.426 & \textbf{0.401}  & 0.427 & \underline{0.445} & 0.431 & 0.446 & 0.445 & 0.460  & \underline{0.420} & 0.440 & 0.831 & 0.657 \\ \hline
        
        \parbox[t]{2mm}{\multirow{4}{*}{\rotatebox[origin=c]{90}{ETTm1}}} & 96 & \textbf{0.298} & \textbf{0.355} & 0.312 & 0.371 & \underline{0.311} & 0.380 & 0.357 & 0.381 & 0.317 & 0.358  & 0.334 & 0.368 & 0.329 & \underline{0.367} & 0.349 & 0.381  & 0.355 & 0.376 & 0.345 & 0.372 \\
        & 192 & \textbf{0.347} & \textbf{0.381} & \underline{0.365} & 0.409 & 0.389 & 0.419 & 0.451 & 0.466 & 0.361 & 0.387  & 0.377 & 0.391 & 0.367 & \underline{0.385} & 0.385 & 0.390  & 0.391 & 0.392 & 0.380 & 0.389 \\
        & 336 & \textbf{0.374} & \underline{0.397} & 0.421 & 0.410 & 0.401 & 0.417 & 0.483 & 0.511 & 0.383 & \textbf{0.378}  & 0.426 & 0.420 & \underline{0.399} & \underline{0.410} & 0.415 & 0.402  & 0.424 & 0.415 & 0.413 & 0.413 \\
        & 720 & \textbf{0.429} & \textbf{0.428} & 0.496 & \underline{0.437} & 0.488 & 0.476 & 0.485 & 0.507 & \underline{0.437} & 0.439  & 0.491 & 0.459 & 0.454 & 0.439 & 0.487 & 0.467  & 0.487 & 0.450 & 0.474 & 0.453 \\ \hline
        
        \parbox[t]{2mm}{\multirow{4}{*}{\rotatebox[origin=c]{90}{ETTm2}}} & 96 & \underline{0.176} & \underline{0.264} & 0.185 & 0.290 & 0.191 & 0.301 &  0.251 & 0.332 & 0.199 & 0.288  & 0.180 & \underline{0.264} & \textbf{0.175} & \textbf{0.259} & 0.205 & 0.281  & 0.182 & 0.265 & 0.193 & 0.292 \\
        & 192 & \textbf{0.231} & \underline{0.303} & 0.292 & 0.309 & 0.253 & 0.312 & 0.287 & 0.312 & 0.257 & 0.328  & 0.250 & 0.309 & \underline{0.241} & \textbf{0.302} & 0.257 & 0.312  & 0.246 & 0.304 & 0.284 & 0.362 \\
        & 336 & \textbf{0.290} & \underline{0.339} & 0.321 & 0.367 & \underline{0.298} & 0.342 & 0.303 & \textbf{0.337} & 0.309 & 0.363  & 0.311 & 0.348 & 0.305 & 0.343 & 0.310 & 0.388  & 0.307 & 0.342 & 0.369 & 0.427 \\
        & 720 & \textbf{0.388} & \textbf{0.398} & 0.401 & \underline{0.400} & 0.409 & 0.407 & 0.420 & 0.431 & \underline{0.395} & 0.416  & 0.412 & 0.407 & 0.402 & \underline{0.400} & 0.401 & 0.414  & 0.407 & \textbf{0.398} & 0.554 & 0.522 \\ \hline
        
        \parbox[t]{2mm}{\multirow{4}{*}{\rotatebox[origin=c]{90}{Weather}}} & 96 & \textbf{0.153} & \underline{0.205} & 0.174 & 0.218 & \underline{0.169} & 0.221 & 0.187 & \textbf{0.192} & 0.181 & 0.239  & 0.174 & 0.214 & 0.177 & 0.218 & 0.190 & 0.239  & 0.192 & 0.232 & 0.196 & 0.255 \\
        & 192 & \textbf{0.196} & \underline{0.244} & \underline{0.200} & 0.258 & 0.205 & \underline{0.248} & 0.212 & 0.265 & 0.238 & 0.272  & 0.221 & 0.254 & 0.225 & 0.259 & 0.212 & 0.270  & \textbf{0.240} & 0.271 & 0.237 & 0.296 \\
        & 336 & \textbf{0.246} & \textbf{0.283} & 0.280 & 0.299 & 0.288 & 0.299 & 0.272 & \underline{0.288} & 0.285 & 0.315  & 0.278 & 0.296 & 0.278 & 0.297 & \underline{0.271} & 0.308  & 0.292 & 0.307 & 0.283 & 0.335 \\
        & 720 & \textbf{0.314} & \textbf{0.334} & 0.352 & 0.359 & \underline{0.335} & 0.369 & 0.360 & 0.369 & 0.352 & 0.377  & 0.358 & \underline{0.347} & 0.354 & 0.348 & 0.358 & 0.363  & 0.364 & 0.353 & 0.345 & 0.381 \\ \hline
        
        \parbox[t]{2mm}{\multirow{4}{*}{\rotatebox[origin=c]{90}{ECL}}} & 96 & \textbf{0.141} & \textbf{0.239} & 0.156 & \underline{0.240} & 0.157 & 0.255 & 0.167 & 0.261 & 0.171 & 0.264  & \underline{0.148} & \underline{0.240} & 0.181 & 0.270 & 0.181 & 0.275  & 0.201 & 0.281 & 0.197 & 0.282 \\
        & 192 & \textbf{0.159} & 0.255 & 0.161 & 0.268 & 0.188 & 0.271 & 0.187 & \textbf{0.253} & 0.188 & 0.281  & 0.162 & \textbf{0.253} & 0.188 & 0.274& \underline{0.160} & \underline{0.254} & 0.201 & 0.283 & 0.196 & 0.285 \\
        & 336 & \textbf{0.171} & \textbf{0.268} & 0.195 & 0.272 & 0.192 & 0.275 & 0.199  & 0.284 & 0.207 & 0.298  & \underline{0.178} & \underline{0.269} & 0.204 & 0.293 & 0.191 & 0.286  & 0.215 & 0.298 & 0.209 & 0.301 \\
        & 720 & \textbf{0.208} & \textbf{0.300} & 0.231 & \underline{0.307} & 0.241 & 0.339 & 0.250 & 0.314 & 0.244 & 0.333  & \underline{0.225} & 0.317 & 0.246 & 0.324 & 0.247 & 0.318  & 0.257 & 0.331 & 0.245 & 0.333 \\ \hline
        
        \parbox[t]{2mm}{\multirow{4}{*}{\rotatebox[origin=c]{90}{Traffic}}} & 96 & \textbf{0.367} & \textbf{0.257} & \underline{0.395} & 0.268 & 0.401 & \underline{0.259} & 0.511 & 0.342 & 0.392 & 0.276  & \underline{0.395} & 0.268 & 0.462 & 0.295 & 0.412 & 0.288  & 0.649 & 0.389 & 0.650 & 0.396 \\
        & 192 & \underline{0.385} & \textbf{0.266} & 0.393 & 0.282 & 0.389 & 0.294 & 0.492 & 0.377 & 0.419 & 0.297  & 0.417 & \underline{0.276} & \textbf{0.366} & 0.296 & 0.468 & 0.304  & 0.601 & 0.366 & 0.598 & 0.370 \\
        & 336 & \textbf{0.401} & \textbf{0.275} & 0.443 & 0.368 & \underline{0.427} & 0.296 & 0.530  & 0.351 & 0.434 & 0.304  & 0.433 & \underline{0.283} & 0.482 & 0.304 & 0.436 & 0.304  & 0.609 & 0.369 & 0.605 & 0.373 \\
        & 720 & \textbf{0.445} & \textbf{0.302} & 0.470 & 0.309 & 0.473 & 0.347 & 0.547 & 0.367 & 0.471 & 0.324  & \underline{0.467} & \textbf{0.302} & 0.514 & 0.322 & 0.474 & \underline{0.307} & 0.647 & 0.387 & 0.645 & 0.394 \\
    \bottomrule
    \end{tabular}\label{tab:reordered_results}
    }
\end{table*}

\section{Experiments}\label{sec:exp}
In this section, we describe the details of experiments, including experimental setup, time series forecasting results, ablation studies, and analysis on memory and speed.

\subsection{Experimental Setup}
\textbf{Datasets.} To evaluate the proposed {\m}, we adopt seven popular real-world datasets~\cite{liu2024iTransformer}, including ETTh1\&ETTh2, ETTm1\&ETTm2, Weather, ECL, and Traffic. The description of these multivariate time series datasets are summarized in Table~\ref{tab:dataset}.
\\\textbf{Baselines and Metrics.} We compare SST with representative time series forecasting methods within three years. They include Mamba-based models (TimeMachine~\cite{ahamed2024timemachine}, S-Mamba~\cite{wang2024mamba}, and SiMBA~\cite{patro2024simba}), Transformer-based forecasters (MTST~\cite{zhang2024multi}, iTransformer~\cite{liu2024iTransformer}, and PatchTST~\cite{nie2023a}), MLP-based framework (TimeMixer~\cite{wang2024timemixer}, RLinear~\cite{li2023revisiting}, and DLinear~\cite{zeng2023transformers}) .
Note that it is unfair to include LLMs or large time series foundation models as they are pre-trained on a large corpus of data. To assess the performance of time series forecasters, we adopt two widely-used metrics MSE and MAE. The lower MSE and MAE indicate more accurate forecasting results.
\\\textbf{Experimental Setting.} We use low-resolution $R_{PTS}=0.43$ ($P_L=48$ and $Str_L=16$) for long-range time series and high-resolution $R_{PTS}=0.5$ ($P_L=16$ and $Str_L=8$) for short range. 
We set the look-back window length $L=2S=672$, and the future value length $F\in \{96, 192, 336, 720\}$. We run all experiments (including baselines) ten times with different seeds and report their average performance.
\subsection{Time Series Forecasting Results.} 
Table~\ref{tab:reordered_results} presents the main time series forecasting results. Overall, {\m} consistently outperforms baseline models, including Mamba-based, Transformer-based, and MLP-based approaches across all real-world datasets. For instance, on the ETTm1 dataset with the longest forecast horizon ($F=720$), {\m} achieves an MSE of 0.429, compared to 0.488 for S-Mamba, 0.491 for iTransformer, and 0.487 for TimeMixer, reflecting relative improvements of 13.75\%, 14.45\%, and 13.51\%, respectively. This superior performance highlights the effectiveness of {\m}'s architectural design. By leveraging low-resolution, long-range representations, {\m} employs Mamba to extract global patterns, while using high-resolution, short-range representations allows the Transformer-based LWT module to capture local variations. A subsequent long-short router adaptively integrates long- and short-range dependencies, enabling more accurate forecasting. Notably, {\m} outperforms MTST~\cite{zhang2024multi} which employs Transformer to model diverse temporal patterns at different resolutions in multiple branches. This comparison suggests that modeling multi-scale resolution alone may not be the most effective method for time series forecasting. 

\subsection{Ablation Studies}
\begin{figure}[t]
    \centering
    \subfigure[]{
    	\begin{minipage}{0.22\textwidth}
   		 	\includegraphics[width=1\textwidth]{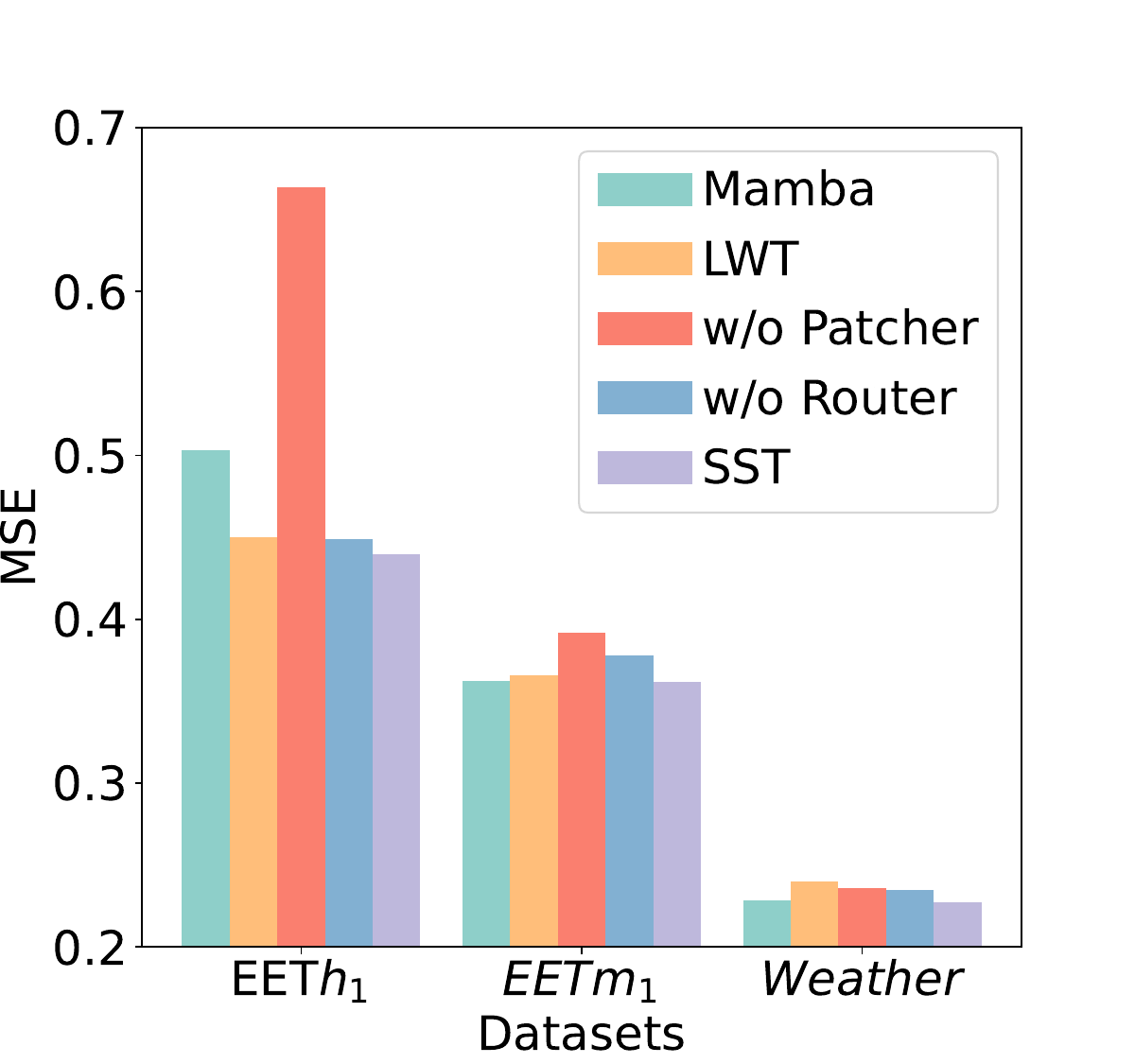}
    	\end{minipage}
    }
    \subfigure[]{
    	\begin{minipage}{0.22\textwidth}
   		 	\includegraphics[width=1\textwidth]{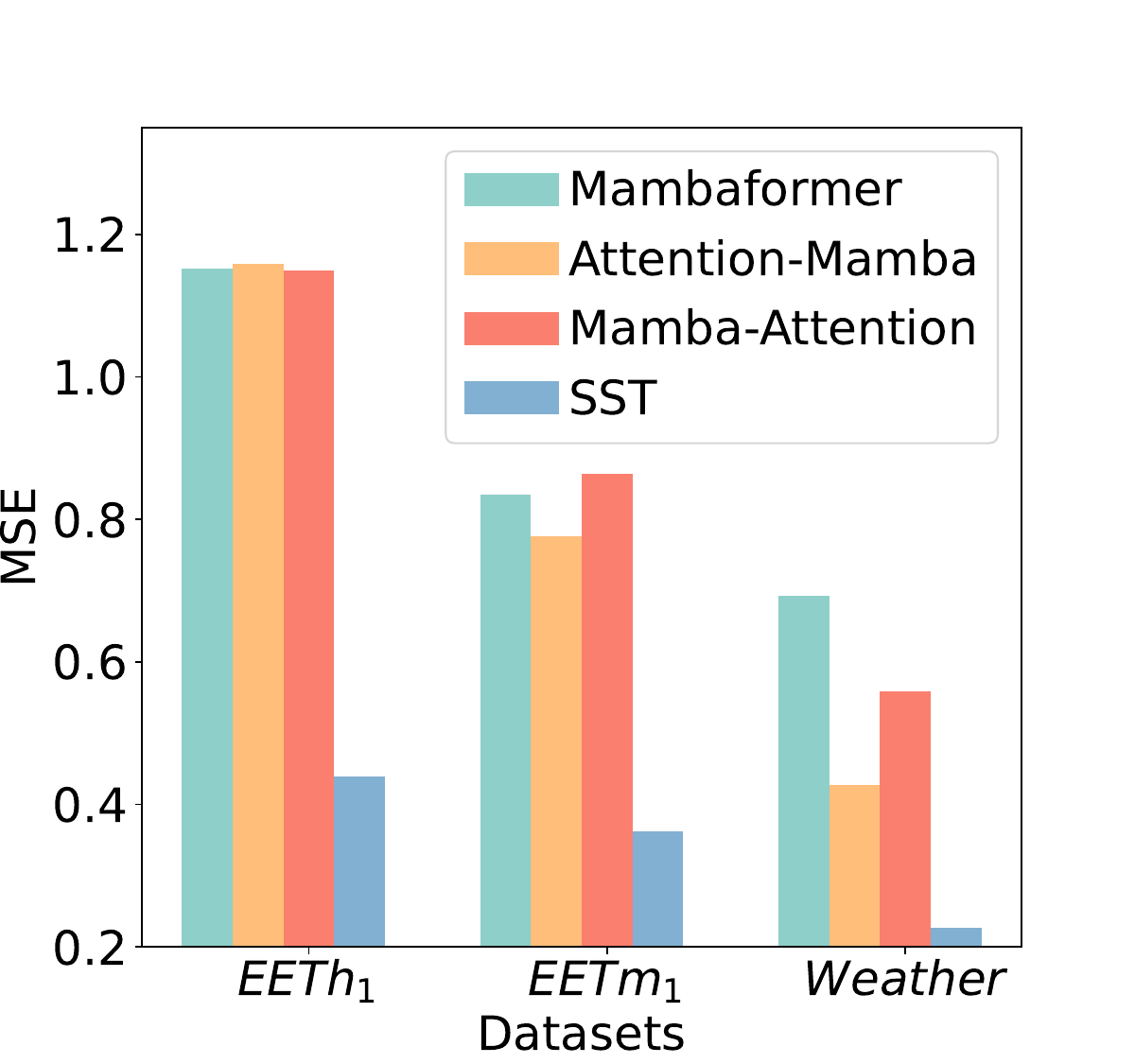}
    	\end{minipage}
    }
    \vspace{-0.4cm}
    \caption{Ablation study on three datasets for (a) exclusion of key components; (b) comparison with Mambaformer family.}
    \label{fig:ablation}
\end{figure}

To assess the contribution of each component in {\m}, we conduct ablation studies on the following variants: \textit{Mamba}, \textit{LWT}, \textit{w/o Patcher} (SST without the patching module), and \textit{w/o Router} (SST without the routing module), and also comparing {\m} with Mambaformer family. Results are presented in Figure~\ref{fig:ablation}, and we draw the following observations:
(1) {\m} significantly outperforms both \textit{Mamba} and \textit{LWT}, highlighting the advantage of combining Mamba and Transformer. This hybrid design enables {\m} to effectively capture long-range patterns and short-range variations, validating the benefit of integrating state space and attention mechanisms.
(2) {\m} also surpasses \textit{w/o Patcher} and \textit{w/o Router}, demonstrating the importance of the multi-scale patcher and dynamic router. The patcher facilitates resolution-aware processing by adapting time series granularity for long- and short-range segments, while the router dynamically balances contributions from the two expert modules. These results confirm the effectiveness of {\m}'s multi-scale design and MoE-based hybrid architecture. (3) Compared to the {\mf} family, {\m} achieves significantly better performance, further emphasizing the benefits of its carefully integrated hybrid components.

\subsection{Memory and Speed Analysis}
\begin{figure}[t]
    \centering
    \subfigure[Memory]{
    	\begin{minipage}{0.22\textwidth}
   		 	\includegraphics[width=1\textwidth]{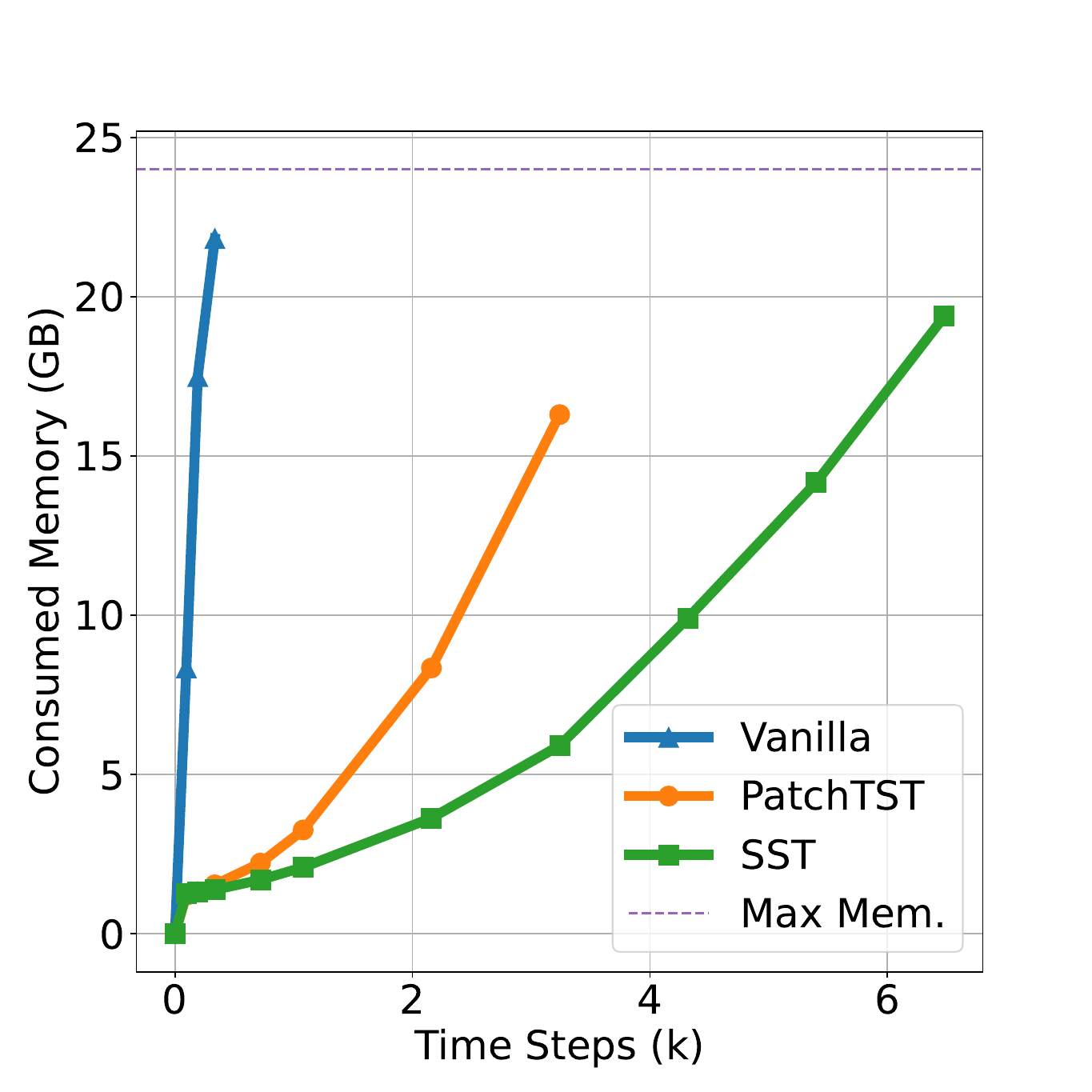}
    	\end{minipage}
            \label{fig:memory}
    }
    \subfigure[Time]{
    	\begin{minipage}{0.22\textwidth}
   		 	\includegraphics[width=1\textwidth]{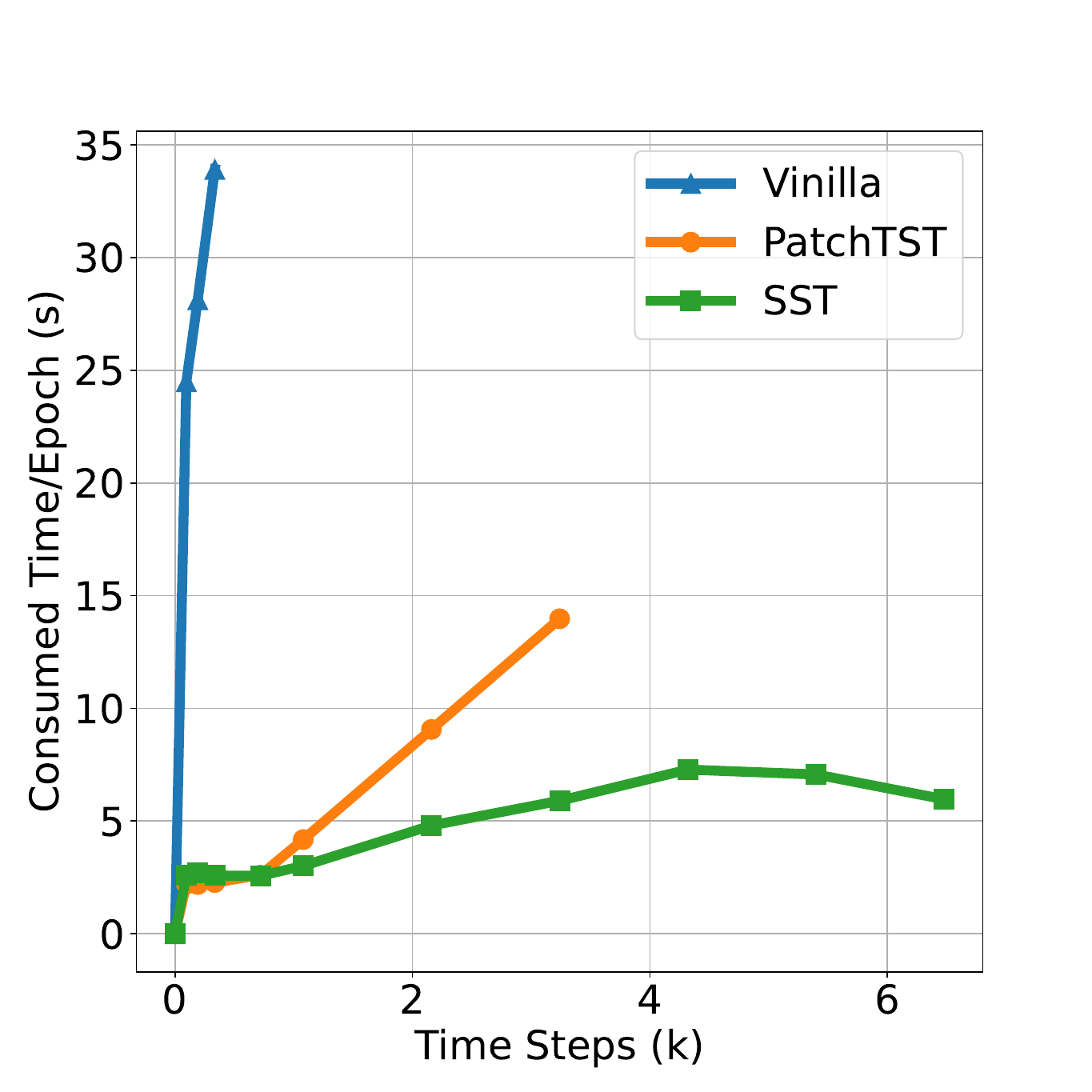}
    	\end{minipage}
            \label{fig:time}
    }
    \vspace{-0.4cm}
    \caption{Comparison of consumed memory and the elapsed time of each epoch for vanilla Transformer, PatchTST, and {\m}. The OOM (out-of-memory) issues emerge for vanilla transformer, PatchTST, and {\m} when the length of input is 336, 3240, and 6480 time steps. It demonstrates the efficiency of {\m} in terms of both memory and time.}
    \vspace{-0.7cm}
    \label{tab:memtime}
\end{figure}

To check the efficiency of {\m} in practice, we conduct memory and speed analysis. As shown in Figure~\ref{tab:memtime}, we depict figures where consumed memory and the elapsed time each epoch vary with the length of input time series $L$. We compare {\m} with PatchTST~\cite{nie2023a} and vanilla Transformer~\cite{vaswani2017attention}. Note that we do not compare iTransformer because its attention operates on variate dimension instead of sequence dimension. All the computations were performed on 24 GB NVIDIA RTX A5000 GPU at Ubuntu 20.04.6 LTS. From the figure, we observe {\m} is efficient and has promising scalability on time steps. {\m} is able to scale linearly to 6k time steps. The impressive scalability stems from the fact that Mamba implement a hardware-aware algorithm, and LWT leverages a fixed-size sliding window to operate attentions. In contrast, vanilla Transformer struggles in quadratic complexity, significantly preventing it from attending to long time series. In the experiments, vanilla Transformer can only attend to maximum 336 time steps. It cannot continue increasing input length due to the OOM (Out-Of-Memory) issue. PatchTST uses patch techniques to reduce complexity by a factor of stride, thus alleviating the issue to a degree. However, it still fall short of scalability and can only scale to 3k time steps, much lower than 6k time steps of {\m}. The memory and speed analysis demonstrate efficiency of {\m} in practice.

\section{Related Work}\label{sec:related}
This work is primarily related to two lines of research: (1) time series forecasting and (2) state space models.

\subsection{Time Series Forecasting}
Time series forecasting has long been a fundamental problem. Classical deep learning approaches such as RNNs~\cite{cheng2018deeptransport, hewamalage2021recurrent} and LSTMs~\cite{yao2019revisiting, xu2024surrogate} have been widely applied, but their effectiveness on long-range sequences is limited due to gradient vanishing issues~\cite{tetko1995neural}. Inspired by the success of Transformers~\cite{vaswani2017attention} in natural language processing, numerous Transformer-based models have been adapted for time series forecasting~\cite{ijcai2023p759, li2019enhancing, zhou2021informer, liu2021pyraformer, wu2021autoformer, zhou2022fedformer, nie2023a}. For example, iTransformer~\cite{liu2024iTransformer} achieves state-of-the-art performance by applying attention and feed-forward networks on the inverted dimensions. More recently, the powerful large language models (LLMs) have been explored for time series forecasting~\cite{jin2023time, gruver2024large, tan2024language}. However, these approaches remain computationally expensive due to their super-linear complexity and a huge amount of parameters contained in the model. In parallel, the ideas of multi-scale time series has been increasingly adopted. MTST~\cite{zhang2024multi} and TimeMixer(++) extract multi-scale features for predicative analysis. Although they also consider different time series granularities, they do not design different models for specific granularities. Different from them, our work employs different specifically designed models to capture patterns in long range and fluctuations in short range.

\subsection{State Space Models}
SSMs~\cite{gu2021combining, gu2022efficiently, gu2023mamba} have recently emerged as a powerful family of sequence modeling architectures. Mamba, a selective SSM equipped with a hardware-efficient algorithm, has demonstrated impressive performance across a wide range of modalities, including language~\cite{gu2023mamba, dao2024transformers}, vision~\cite{zhu2024vision, tang2024vmrnn, yu2024mambaout}, medicine~\cite{ma2024u, xing2024segmamba}, graphs~\cite{wang2024graph, behrouz2024graph}, recommendation systems~\cite{liu2024mamba4rec, yang2024uncovering}, and time series~\cite{ahamed2024timemachine, wang2024mamba, patro2024simba, liang2024bi}. Recent efforts have proposed leveraging SSMs for time series forecasting~\cite{ahamed2024timemachine, wang2024mamba, patro2024simba}. For instance, TimeMachine~\cite{ahamed2024timemachine} uses stacked Mamba blocks to model long-range dependencies in multivariate time series. However, these models are limited in capturing fine-grained variations because of the inherent lossy memory mechanism. In contrast, our work makes the first attempt to integrate Mamba and Transformer within a unified architecture for time series forecasting, aiming to capture both long- and short-range temporal dynamics, and also to eliminate \textit{information interference} issue in the straightforward stacking of Mamba and Transformer.

A notable research direction involves hybridizing Mamba with Transformers. For example, Mambaformer~\cite{park2024can} demonstrates strong performance in in-context learning tasks by interleaving Mamba and attention layers. Jamba~\cite{lieber2024jamba} is a production-scale hybrid model that combines attention and SSMs, supporting up to 52B parameters and long-context modeling. While these hybrid approaches have focused on language modeling, our work is the first to explore a hybrid Mamba–Transformer architecture for time series forecasting, leveraging their complementary strengths to handle both long-term dependencies and short-term variations.

\section{Conclusion and Future Work}
In this paper, we analyze the strengths and weaknesses of Mamba and Transformer architectures: Mamba offers linear complexity and computational efficiency but may struggle with information retention, while Transformer provides direct access to past information at the cost of quadratic complexity. Motivated by this, we raise a research question: \textit{Can we design a hybrid Mamba-Transformer architecture that is both effective and efficient for time series forecasting?} To explore this, we adapt {\mf} to time series forecasting, which is the hybrid Mamba-Transformer architecture by stacking Mamba and attention layers for language modeling. The results show that such integration is ineffective for time series due to the \textit{information interference} issue. To address the issue, our proposed {\m} leverages Mamba to extract long-term patterns in coarse-grained time series and Transformer to capture short-term variations in fine-grained time series. Extensive experiments demonstrate that our approach achieves state-of-the-art (SOTA) performance while maintaining linear complexity $O(L) $. Future work explores the application of hybrid Mamba-Transformer architectures to other time series analysis tasks such as classification and anomaly detection.

\begin{acks}
This material is based upon work supported by NSF awards (SaTC-2241068, IIS-2506643, and POSE-2346158), a Cisco Research Award, and a Microsoft Accelerate Foundation Models Research Award.
\end{acks}

\section{GenAI Usage Disclosure}
This paper presents original work. After completing the manuscript, we used the GenAI tool ChatGPT-4o to assist with polishing the abstract and introduction.

\bibliographystyle{ACM-Reference-Format}
\bibliography{sample-base}

\end{document}